\documentclass{article}

\usepackage[preprint]{neurips_2026}

\usepackage[utf8]{inputenc} 
\usepackage[T1]{fontenc}    
\usepackage{hyperref}       
\usepackage{url}            
\usepackage{booktabs}       
\usepackage{amsfonts}       
\usepackage{nicefrac}       
\usepackage{microtype}      
\usepackage{xcolor}         
\usepackage{enumitem}
\usepackage{amsmath}
\usepackage{amsthm}
\usepackage{graphicx}
\usepackage{wrapfig}
\usepackage{algorithm}
\usepackage{algpseudocode}

\title{Enforcing Constraints in Generative Sampling via Adaptive Correction Scheduling }

\author{%
    Noah Trupin \qquad Yexiang Xue \\
    Department of Computer Science \\
    Purdue University \\
    \texttt{\{ntrupin,yexiang\}@purdue.edu}
}

\begin{document}

\maketitle

\begin{abstract}
    Hard constraints in generative sampling are typically enforced by projection, applied either once at the end of sampling or after every update. This binary framing overlooks a fundamental issue: \emph{projection changes the distribution of states which future updates depend on.} As a result, delayed projection can produce samples that are feasible but inconsistent with the intended sampling dynamics, even after final projection. We formalize constraint enforcement as a correction scheduling problem over the generative rollout. Using one-step constraint defect as a local signal of geometric mismatch, we introduce \emph{adaptive correction scheduling}, a state-dependent policy that allocates projection budget to the steps that most strongly perturb the trajectory. Terminal and stepwise projection arise as limiting cases of this family. Across controlled manifold rollouts and a learned projected diffusion sampler, adaptive scheduling improves the cost--accuracy frontier at matched projection budgets, recovering $71.2\%$ of full stepwise benefit with $75\%$ fewer corrections. These results show that constraint timing is a first-class design variable in generative sampling, and that enforcing feasibility alone is insufficient to preserve the intended constrained sampling dynamics.
\end{abstract}

\begin{figure}[hb!]
    \centering
    \includegraphics[width=\linewidth]{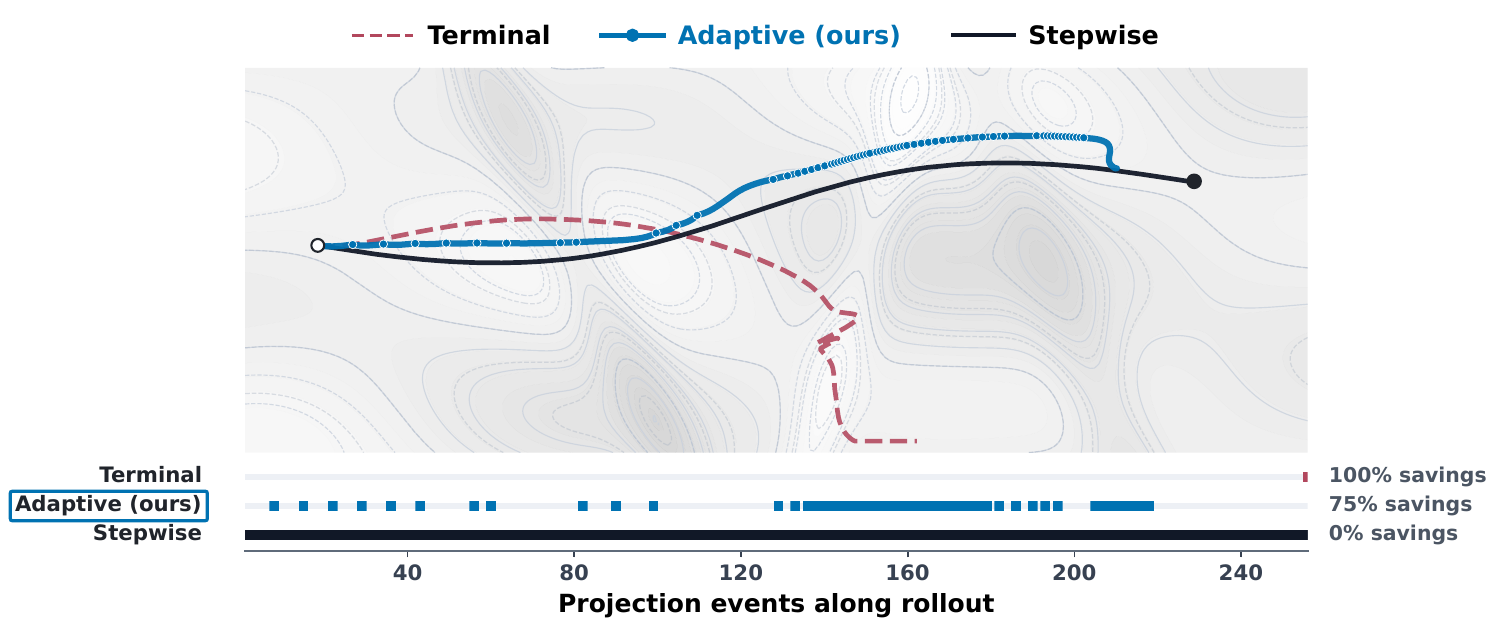}
    \vspace{-12pt}
    \caption{\emph{Adaptive correction preserves rollouts at a fraction of the cost.} Terminal correction projects only after the trajectory has already drifted from the ridged terrain, producing a feasible but dynamically inconsistent path. Stepwise correction prevents drift by projecting after every update, but pays the full projection cost. Our adaptive scheduler uses the same projection operator but spends only $25\%$ of the projection calls, concentrating them where defect is largest to closely track the stepwise rollout.}
    \label{fig:title}
\end{figure}

\section{Introduction}
\label{sec:introduction}

Generative models are increasingly deployed in settings where outputs must satisfy hard constraints, including trajectory synthesis, robot planning, and structured generation~\citep{liDiffuSolveDiffusionbasedSolver2024, liangSimultaneousMultiRobotMotion2025, zhangConstrainedDiffusersSafe2025, christopherConstrainedSynthesisProjected2024, cardeiConstrainedMolecularGeneration2025, santiVerifierConstrainedFlowExpansion2025}. In these domains, samples are required to lie on manifolds defined by geometric, physical, or task-specific structures. A common strategy is to run the generative process in the ambient space and enforce feasibility through projection. This raises the question: \emph{when should projection be applied?}

In practice, two strategies dominate: terminal correction projects only at the end of sampling, while stepwise correction projects after every update. Both are widely used, and both can produce feasible outputs. However, they treat correction timing as a fixed design choice rather than part of the algorithm.

This framing ignores that generative updates are state-dependent, as each step is computed from the current iterate. Once a rollout leaves the constraint set, subsequent updates are evaluated at states that a constrained trajectory would never visit. A later projection can restore feasibility, but it cannot in general undo the distributional shift induced by these off-manifold states. As a result, terminally corrected samples are often feasible but wrong. They lie on the constraint set, but correspond to a trajectory that deviated from the constrained dynamics during rollout.

Constraint enforcement is therefore not only about satisfying feasibility at the endpoint, but about preserving the trajectory distribution that produces that endpoint. This makes correction timing a resource allocation problem. In typical rollouts, most updates remain close to the constraint set, while a smaller subset produces large excursions that drive downstream error. Terminal correction under-allocates effort to these critical steps, while stepwise correction over-allocates effort everywhere. The central challenge is thus how to allocate a limited number of corrections across the rollout.

We address this by reframing constraint enforcement as a correction scheduling problem. We use one-step constraint defect, the distance of a proposed update from the feasible set, as a local signal of geometric mismatch. Small defect indicates that the rollout remains close to the constrained trajectory, while large defect identifies steps whose effects are unlikely to be corrected later. This leads to an online budgeted policy: spend corrections on high-defect steps, where projection is expected to most reduce downstream trajectory error.

This yields a family of online budgeted schedules that interpolate between terminal and stepwise correction while adapting to the realized defect profile of each rollout. Fundamentally, this policy is not a heuristic layered onto existing methods, but a direct consequence of viewing constraint enforcement through the lens of trajectory consistency. It is model-agnostic, requires no gradients through projection, and introduces negligible overhead beyond defect evaluation.

Our contribution is a scheduling view of constrained generative sampling: \begin{enumerate}[label=\arabic*.]
    \item \emph{Timing changes the rollout.} We show that delayed correction alters the distribution of future updates, producing endpoint deviations that cannot be resolved by terminal projection alone.

    \item \emph{Constraint enforcement is a budget allocation problem.} We formalize correction as allocating a limited number of projections across rollout, making terminal, periodic, and adaptive strategies directly comparable.

    \item \emph{Defect provides an effective value proxy.} An online budgeted scheduler uses defect to spend corrections on the steps that most strongly perturb the trajectory.

    \item \emph{Adaptive scheduling improves the cost–accuracy frontier.} At matched correction cost, adaptive policies recover near-stepwise fidelity while using substantially fewer projections.
\end{enumerate}

Across controlled manifold rollouts and learned diffusion settings, we observe that constraint violations are highly heterogeneous across time, and that selectively correcting high-defect steps consistently outperforms uniform strategies. These results demonstrate that constraint timing is a fundamental degree of freedom in generative sampling, and that enforcing feasibility alone is insufficient to preserve the intended constrained distribution.

\section{Related Work}

Prior work enforces hard constraints in generative modeling by incorporating projection or optimization into the sampling loop~\citep{christopherConstrainedSynthesisProjected2024, santiVerifierConstrainedFlowExpansion2025, utkarshPhysicsConstrainedFlowMatching2025, zhangConstrainedDiffusersSafe2025, hoseinpourConstrainedDiffusionModels2025}. In diffusion and flow-based models, projected sampling methods often apply projection after each step to maintain feasibility throughout generation~\citep{chiDiffusionPolicyVisuomotor2024, liDiffuSolveDiffusionbasedSolver2024, xiaoSafeDiffuserSafePlanning2024, jannerPlanningDiffusionFlexible2022, liangSimultaneousMultiRobotMotion2025, yangUniConFlowUnifiedConstrained2026, santiVerifierConstrainedFlowExpansion2025, utkarshPhysicsConstrainedFlowMatching2025, niPhysicsinformedNeuralMotion2024}. Variants introduce projected gradients, differentiable optimization layers, or discrete constraint operators~\citep{cardeiConstrainedDiscreteDiffusion2025, cardeiConstrainedMolecularGeneration2025, chengGradientFreeGenerationHardConstrained2024, hoseinpourConstrainedDiffusionModels2025}.

These approaches address how to enforce constraints, but largely treat correction timing as fixed. Stepwise projection is typically adopted as a default, while terminal projection is used as a cheaper alternative. Our work instead focuses on when correction should be applied, holding the underlying generative process fixed.

This question is closely related to geometric integration, where the interaction between an update rule and a projection or retraction determines long-horizon behavior~\citep{hairerGeometricIntegrationOrdinary2001, seguinLowrankRetractionsDynamical2024, christopherConstrainedSynthesisProjected2024, mclachlanGeometricGeneralisationsSHAKE2014, linanRetractionMapsSeed2023}. Classical results show that projection and dynamics do not generally commute, leading to trajectory drift even when endpoints are feasible. Our setting differs in two key ways: the dynamics are learned and stochastic, and the objective is  efficient allocation of correction compute alongside accuracy.

A separate line of work studies adaptive schedules in diffusion, selecting timesteps or noise levels to trade off speed and sample quality. Methods such as AdaDiff, TDPM, stepsize distillation, and active noise estimation allocate denoising compute across time~\citep{zhangAdaDiffAdaptiveStep2024, yeScheduleFlyDiffusion2025, peiOptimalStepsizeDiffusion2025, kimModelAlreadyKnows2026}. These approaches operate along an orthogonal axis: they control the generative dynamics themselves. In contrast, we allocate constraint enforcement, treating projection as a limited resource applied to preserve trajectory consistency.

Closest are methods that interleave optimization with sampling, but these apply correction uniformly or through fixed heuristics. We instead treat correction as a budgeted allocation problem.

\section{Adaptive Correction Scheduling}
\label{sec:acs}

Projection timing is a control decision. Rather than projecting only at the end or after every update, we treat correction as a finite resource to be allocated across the rollout. Given a rollout of length $T$ and a correction budget $B$, the scheduler must choose which proposed updates to project. The ideal policy would spend corrections where they most reduce future trajectory error, but this marginal value is not directly observable in practice. Our method uses one-step constraint defect as an online proxy.

At each step, the scheduler first applies the generative update without correction, measures how far the proposed state departs from the constraint set, and then decides whether this departure is worth spending one unit of budget. Large defect indicates that the rollout has entered a region where future updates are likely to differ from the constrained trajectory, while small defect indicates that projection can be safely delayed. As such, the adaptive scheduler spends corrections where the realized rollout is most likely to drift, while matching the projection budget of fixed periodic baselines.

This section formalizes that idea: we describe rollouts with correction events, define defect as the scheduling signal, and derive our budget-aware online policy.

\subsection{Rollouts with correction events}

\begin{figure}
    \centering
    \includegraphics[width=\linewidth]{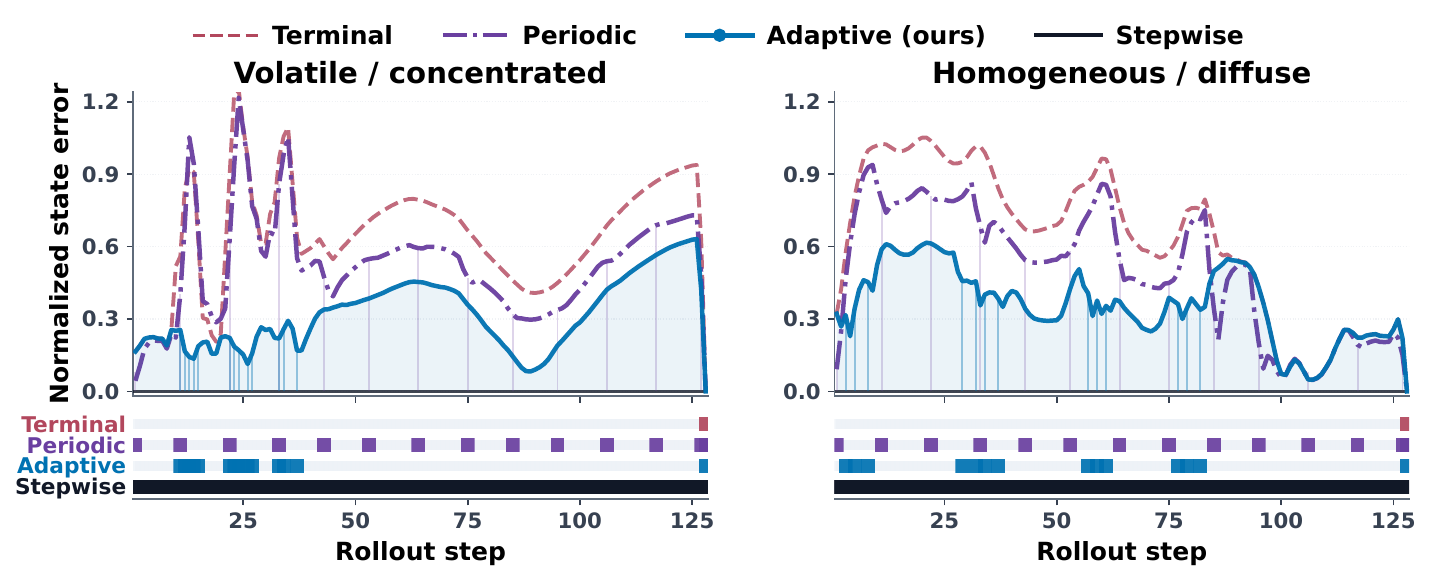}
    \vspace{-8pt}
    \caption{\emph{Adaptive gains appear when trajectory error is concentrated in time.} Each panel shows normalized state error from the stepwise reference alongside projection events. In the volatile terrain setting (left), a few high-defect regions dominate path error; periodic correction spends budget uniformly and misses several of these events, while adaptive correction concentrates projections where they most reduce downstream drift. In the homogeneous setting (right), defects are more diffuse, so the gap between adaptive and periodic narrows. This supports that adaptive correction helps most when a subset of rollout steps controls the trajectory error.}
    \label{fig:regime}
    \vspace{-4pt}
\end{figure}

We consider a generative process that ideally evolves on a constraint set $\mathcal{M} \subset \mathcal{X}$, where $\mathcal{X} \subseteq \mathbb{R}^d$ is an ambient space. In many applications, $\mathcal{M}$ is a smooth embedded manifold or a set defined by hard constraints, such as kinematic feasibility, geometric structure, or physical validity.

In practice, however, generative models operate in the ambient space. A rollout proceeds by repeatedly applying a learned update rule \[
    x_{t+1} = \Phi_h(x_t), \qquad t = 0,\ldots,T-1,
\] where $\Phi_h$ denotes a discrete update operator, such as a reverse diffusion step or flow-matching update, with step size $h$. To enforce feasibility, we assume access to a projection or retraction operator $\Pi : \mathcal{X} \to \mathcal{M}$ which maps arbitrary states back to the constraint set.

The standard objective is to produce a final sample $x_T \in \mathcal{M}$. However, this endpoint view is incomplete, as the generative process is inherently trajectory-dependent: each update is computed from the current state. If a rollout leaves $\mathcal{M}$, subsequent updates are evaluated at states a constrained trajectory would never visit. While later projection may restore feasibility, it generally cannot undo the off-manifold updates already taken. Fig.~\ref{fig:regime} illustrates greater accumulation of error throughout sampling without principled constraint enforcement. As such, projection is an intervention in the dynamics, and applying it changes both the states from which future updates are computed and the final sample.



We formalize this intervention by introducing a correction schedule
$\sigma : \{0,\ldots,T-1\} \to \{0,1\}$, where $\sigma(t)=1$ indicates that projection is
applied after step $t$. Given the proposed update \begin{equation}
    \tilde{x}_{t+1} = \Phi_h(x_t),
    \label{eq:proposed-update}
\end{equation} the corrected rollout is defined by \[
    x_{t+1} = \begin{cases}
        \Pi(\tilde{x}_{t+1}) & \text{if } \sigma(t)=1, \\
        \tilde{x}_{t+1} & \text{if } \sigma(t)=0.
    \end{cases}
\]

This formulation unifies standard strategies: \emph{terminal correction} sets $\sigma(t) = 0$ for all $t < T - 1$ and projects once at the end, \emph{stepwise correction} sets $\sigma(t) = 1$ for all $t$, and \emph{periodic correction} corrects at fixed intervals. All these correspond to different ways of allocating correction events across the rollout. Our goal is to construct a fourth schedule that spends up to the same budget as a periodic rule but chooses correction times from the realized rollout.

\subsection{Defect as the scheduling signal}

To determine when correction is necessary, we require a signal that quantifies how much a proposed update departs from the constraint set. We define a defect function $d(x,\mathcal{M})$ that measures the distance or constraint violation of a state $x$ relative to $\mathcal{M}$. This may be Euclidean distance from the constraint surface or some other domain-specific metric. 

For a proposed update $\tilde{x}_{t+1} = \Phi_h(x_t)$, we define the one-step defect
\begin{equation}
    s_t = d(\tilde{x}_{t+1},\mathcal{M}).
    \label{eq:defect}
\end{equation}

This scalar captures the local mismatch introduced by the update. A small $s_t$ indicates that the rollout remains close to the constraint set, while large $s_t$ implies that the update produces a significant excursion. Large defect marks steps where future updates are likely to differ most from the constrained rollout, making $s_t$ a local proxy for downstream trajectory distortion.

\subsection{Budgeted online allocation}

We distinguish two types of error: \emph{Endpoint error,} the deviation of the final sample from a reference constrained trajectory, and \emph{pathwise error,} cumulative deviation from the constraint set during rollout. 

A final projection eliminates endpoint infeasibility, but not the pathwise error accumulated during rollout. We measure pathwise deviation via cumulative defect: \[
    E_{\mathrm{path}} = \sum_{t=1}^{T} d(x_t,\mathcal{M}).
\]

From this perspective, constraint enforcement becomes a resource allocation problem. Each projection incurs cost, and a schedule determines how a limited number of projections is distributed across time. Terminal and stepwise correction are extreme allocations: one
spends the budget at the end, the other spends it everywhere. If large deviations are concentrated in a small subset of steps, both are inefficient. This motivates adaptive, state-dependent schedules.

A budgeted schedule therefore answers an online marginal question: at time $t$, with $b$ corrections remaining, is the current deviation worth spending one correction, or should the budget be reserved for future steps? Let \[
    V_t(x_t) = \mathbb{E}\!\left[
        E_{\mathrm{uncorrected}} - E_{\mathrm{corrected}}
        \mid x_t
    \right]
\] denote the downstream value of correcting at step $t$. This quantity captures how much correcting the current step changes the remainder of the rollout.

Given a rollout of length $T$ with a finite correction budget $B$, the ideal policy would
allocate corrections to maximize total reduction in trajectory error:
\begin{equation}
    \max_{\sigma} \sum_{t=0}^{T-1} \sigma(t) V_t(x_t)
    \quad \text{s.t.} \quad
    \sum_{t=0}^{T-1} \sigma(t) \le B,
    \label{eq:budget-objective}
\end{equation}
where $\sigma(t) \in \{0,1\}$ indicates whether correction is applied at step $t$.

This formulation makes explicit that correction scheduling is a budgeted allocation problem over time. In practice, however, $V_t$ is not directly observable. Computing it would require estimating how a correction changes the remainder of the rollout under future
stochastic updates. We therefore require a tractable online proxy. Noting that large one-step defect identifies precisely the states where the rollout has entered regions that induce large downstream deviation, we use defect as a surrogate of this marginal value.

\subsection{Budget-aware thresholds}

The ideal policy in Eq.~\eqref{eq:budget-objective} requires the downstream value $V_t(x_t)$ of correcting at each step. This value is not observable online. We therefore use one-step defect $s_t$ as a local proxy for marginal value, but unlike a fixed-threshold rule, the decision must account for the remaining budget.

Let $b_t$ denote the number of corrections remaining before step $t$. Our scheduler uses a family of thresholds \[
    \lambda_{t,b},
    \qquad
    t=0,\ldots,T-1,\quad b=0,\ldots,B,
\] where $\lambda_{t,b}$ is the marginal price of spending one correction at time $t$ with $b$ corrections remaining. Given a proposed update Eq.~\eqref{eq:proposed-update} and
defect Eq.~\eqref{eq:defect}, the online budgeted policy is \[
    x_{t+1} = \begin{cases}
        \Pi(\tilde{x}_{t+1})
        & \text{if } b_t>0 \text{ and } s_t \ge \lambda_{t,b_t}, \\
        \tilde{x}_{t+1}
        & \text{otherwise}.
    \end{cases}
\] If projection is applied, the remaining budget is updated as $b_{t+1}=b_t-1$; otherwise $b_{t+1}=b_t$.

The dependence on both $t$ and $b$ proves essential: the same defect may be worth correcting late in the rollout, when few future opportunities remain, but not early, when the scheduler should
reserve budget for larger future excursions. Thus $\lambda_{t,b}$ implements an online marginal-value threshold conditioned on the remaining horizon and correction budget.

The thresholds $\lambda_{t,b}$ determine how aggressively the scheduler spends its remaining corrections. In principle, they approximate the value boundary of the budgeted allocation
problem: correction is applied when the observed defect is large enough to justify consuming one unit of budget. In Sec.~\ref{sec:experiments}, thresholds are estimated on held-out calibration rollouts. For each time $t$ and remaining budget $b$, we choose $\lambda_{t,b}$ so that the scheduler spends corrections at the desired pace while preserving budget for future high-defect events. This yields an online policy that uses exactly the same total projection budget as periodic baselines, but allocates that budget according to the geometry of the realized rollout. In our experiments, this surface is estimated by held-out quantiles of future defect traces.

This budget-aware policy reduces to familiar strategies as special cases: setting all thresholds to $+\infty$ recovers terminal correction and setting all feasible thresholds to $0$ recovers stepwise correction.

\subsection{Algorithm}

\begin{wrapfigure}{R}{0.55\textwidth}
    \vspace{-24pt}
    \begin{minipage}{\linewidth}
        \begin{algorithm}[H]
            \caption{Online Adaptive Correction Scheduling}
            \label{alg:adaptive-scheduler}
            \begin{algorithmic}[1]
                \Require budget $B$, thresholds $\{\lambda_{t,b}\}_{t=0,b=0}^{T-1,B}$
                \State $b \leftarrow B$
                \For{$t=0,\ldots,T-1$}
                    \State Propose $\tilde{x}_{t+1} \leftarrow \Phi_h(x_t)$
                    \State Compute defect $s_t \leftarrow d(\tilde{x}_{t+1},\mathcal{M})$
                    \If{$b>0$ and $s_t \ge \lambda_{t,b}$}
                        \State $x_{t+1} \leftarrow \Pi(\tilde{x}_{t+1})$
                        \State $b \leftarrow b-1$
                    \Else
                        \State $x_{t+1} \leftarrow \tilde{x}_{t+1}$
                    \EndIf
                \EndFor
            \end{algorithmic}
        \end{algorithm}
    \end{minipage}
    \vspace{-18pt}
\end{wrapfigure}

Alg.~\ref{alg:adaptive-scheduler} summarizes the online budgeted scheduler. The policy observes only the proposed update, its defect, the current time, and the remaining correction budget. It does not require gradients through projection, retraining the generative model, or modifying the update rule. Its overhead is the cost of evaluating the defect and comparing it to the precomputed threshold $\lambda_{t,b}$.

Since the policy conditions on remaining budget, it avoids the main failure mode of fixed thresholding: spending corrections too early on moderate defects and exhausting the budget before larger excursions occur. The scheduler is therefore adaptive in two senses: it responds to the realized geometry of the rollout through $s_t$, and it adapts its spending rule to the remaining budget through $\lambda_{t,b}$.

\section{Experiments}
\label{sec:experiments}

We evaluate adaptive correction scheduling across manifold-valued and trajectory diffusion settings. The experiments ask: (i) how projection timing affects trajectory error, (ii) whether online allocation improves performance at fixed correction budget, and (iii) whether defect concentration explains these gains.

\subsection{Projection timing induces persistent trajectory error}

\begin{figure}
    \centering
    \includegraphics[width=\linewidth]{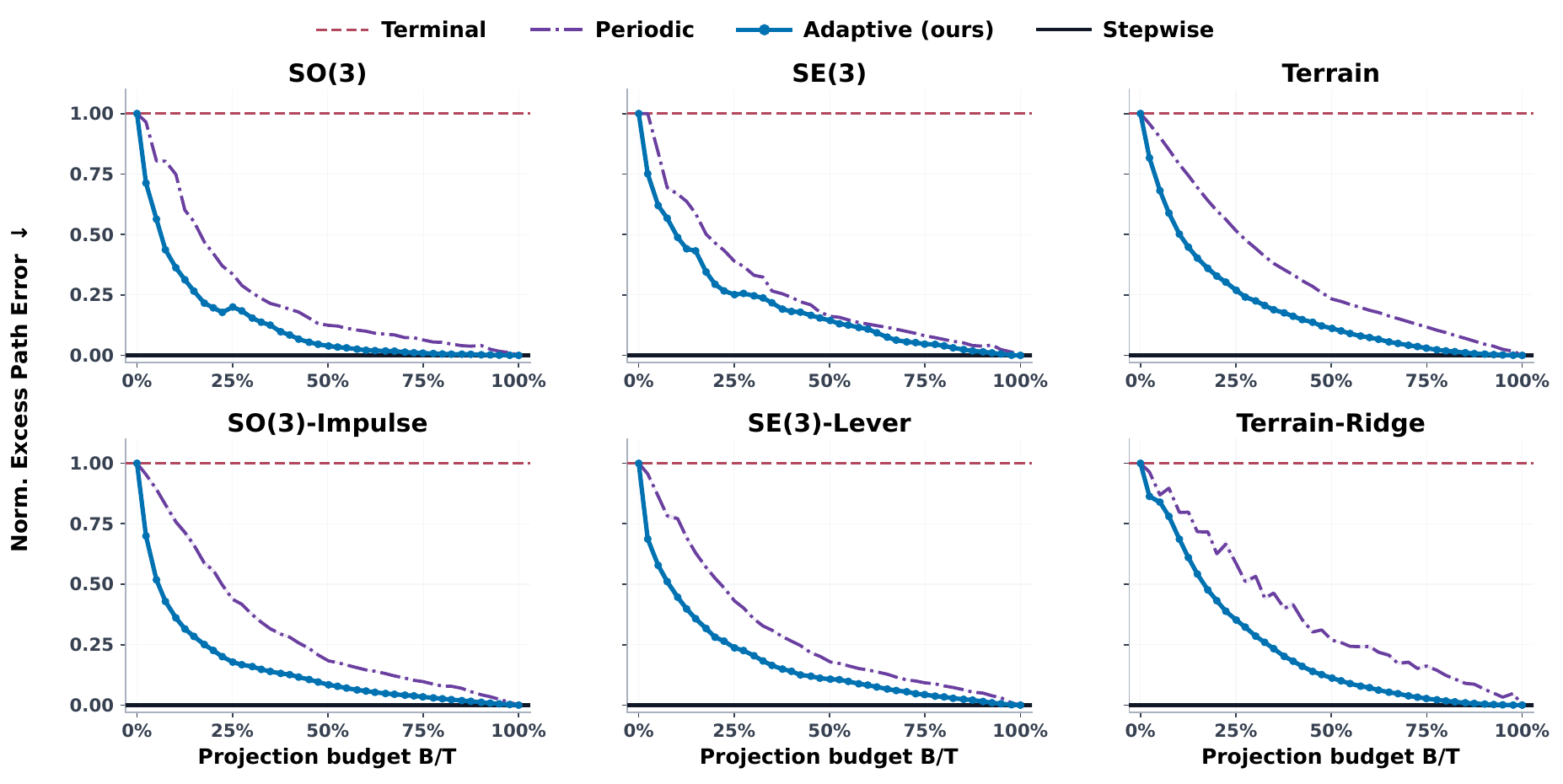}
    \caption{\emph{Adaptive scheduling improves the cost--accuracy frontier across controlled manifolds.} We plot Normalized Excess Path Error (NEPE), where stepwise correction is $0$ and terminal correction is $1$; lower is better. At matched projection budget $B/T$, adaptive correction consistently lies below periodic correction, showing that \emph{where} projections are applied matters more than simply applying them at a fixed frequency. The largest gains occur in the impulse, lever, and ridge variants, where defect is most concentrated.}
    \label{fig:pathwise}
    \vspace{-4pt}
\end{figure}

We begin by isolating the effect of correction timing. Fig.~\ref{fig:title} visualizes representative rollouts under terminal, periodic, and adaptive schedules. Although terminal correction restores feasibility at the final step, the resulting trajectory deviates substantially from the constrained rollout.

This effect is quantified in Tab.~\ref{tab:summary}, which measures endpoint distance to the stepwise reference as a function of correction budget. Even at low defect levels, delayed correction induces a persistent bias: the rollout evolves off-manifold, and subsequent updates compound this deviation. Final projection enforces feasibility but does not recover the original trajectory. An accompanying figure for the endpoint distances reported in Tab.~\ref{tab:summary} can be found in App.~\ref{app:additional-results}.

These results establish that constraint enforcement cannot be understood purely as a final projection step. Projection timing affects the entire rollout, and the resulting error grows continuously with defect.

\subsection{Scheduling dominates frequency at fixed budget}

\begin{table}[b]
    \vspace{-4pt}
    \centering
    \small
    \scalebox{0.86}{\begin{tabular}{lcccccccc}
        \toprule
        & \multicolumn{3}{c}{Endpoint dist. $\downarrow$} & \multicolumn{3}{c}{NEPE $\downarrow$} \\
        \cmidrule(lr){2-4} \cmidrule(lr){5-7}
        Domain & Periodic & Adaptive & $\Delta$ Endpoint $\uparrow$ & Periodic & Adaptive & $\Delta$ NEPE $\uparrow$ \\
        \midrule
        SO(3) & 0.509 $\pm$ 0.0062 & \textbf{0.111 $\pm$ 0.0090} & \textbf{78\% $\pm$ 2\%} & 0.366 $\pm$ 0.0013 & \textbf{0.130 $\pm$ 0.0016} & \textbf{65\% $\pm$ 0\%} \\
        SE(3) & \textbf{0.169 $\pm$ 0.011} & 0.201 $\pm$ 0.0098 & -19\% $\pm$ 7\% & 0.422 $\pm$ 0.0037 & \textbf{0.303 $\pm$ 0.0060} & \textbf{28\% $\pm$ 1\%} \\
        Terrain & \textbf{0.078 $\pm$ 0.017} & 0.083 $\pm$ 0.017 & -6\% $\pm$ 34\% & 0.521 $\pm$ 0.0064 & \textbf{0.323 $\pm$ 0.036} & \textbf{38\% $\pm$ 7\%} \\
        SO(3)-Impulse & 0.030 $\pm$ 0.0035 & \textbf{0.0088 $\pm$ 0.0005} & \textbf{70\% $\pm$ 4\%} & 0.470 $\pm$ 0.015 & \textbf{0.147 $\pm$ 0.015} & \textbf{69\% $\pm$ 4\%} \\
        SE(3)-Lever & 0.064 $\pm$ 0.0066 & \textbf{0.039 $\pm$ 0.0021} & \textbf{40\% $\pm$ 9\%} & 0.486 $\pm$ 0.013 & \textbf{0.252 $\pm$ 0.021} & \textbf{48\% $\pm$ 5\%} \\
        Terrain-Ridge & 0.080 $\pm$ 0.0098 & \textbf{0.079 $\pm$ 0.012} & \textbf{1\% $\pm$ 32\%} & 0.661 $\pm$ 0.039 & \textbf{0.361 $\pm$ 0.021} & \textbf{45\% $\pm$ 4\%} \\
        \bottomrule
    \end{tabular}}
    \vspace{2pt}
    \caption{\emph{Fixed-budget synthetic summary at $B/T \approx 0.25$.} Normalized Excess Path Error (NEPE) is normalized between stepwise correction ($0$) and terminal correction ($1$); lower is better. All methods are evaluated at matched projection budget, and entries report mean $\pm$ SE over paired seeds. Adaptive scheduling consistently improves NEPE, with especially large gains in the concentrated-defect variants; endpoint improvements are geometry-dependent, reflecting that pathwise consistency and final endpoint displacement are related but not identical.}
    \label{tab:summary}
\end{table}

Fig.~\ref{fig:pathwise} shows the main cost--accuracy frontier. Across all six controlled domains, adaptive correction achieves lower NEPE than periodic correction at matched projection budget, with the largest gains in the impulse, lever, and ridge variants. Tab.~\ref{tab:summary} reports the fixed-budget slice at $B/T \approx 0.25$: adaptive consistently improves pathwise fidelity, while endpoint distance is geometry-dependent. This distinction is expected, as pathwise consistency measures whether the rollout followed the constrained dynamics while endpoint displacement also depends on how each geometry maps trajectory errors into final state error.


Paired win-rate diagnostics in App.~\ref{app:additional-results} confirm this trend: adaptive wins on the pathwise metric in 85--90\% of matched synthetic comparisons, showing that at fixed budget, \emph{where} corrections are applied is more important than \emph{how often} they are applied.

\subsection{Adaptive schedules track defect concentration}

\begin{wrapfigure}{r}{0.5\linewidth}
    \vspace{-4pt}
    \centering
    \includegraphics[width=\linewidth]{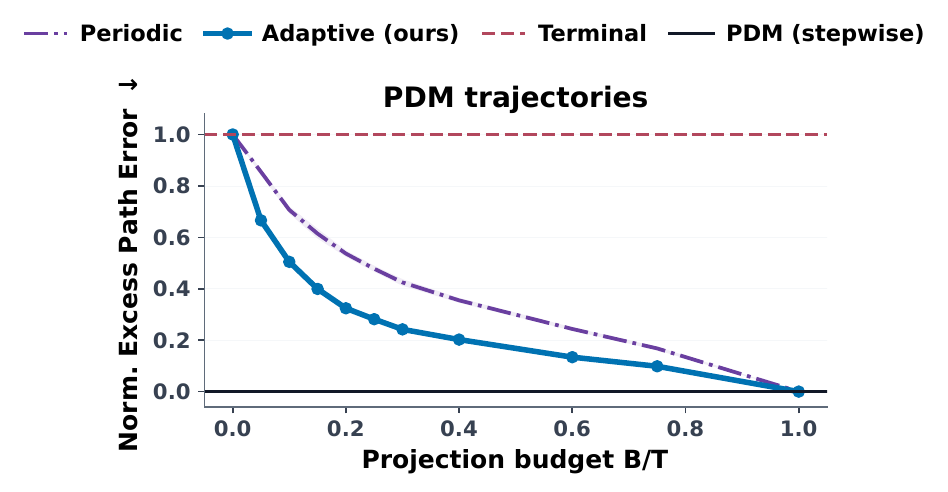}
    \vspace{-8pt}
    \caption{\emph{Adaptive scheduling recovers more of the original PDM sampler at every projection budget.} We keep the PDM model, constraint set, and projection operator fixed, and vary only projection timing. Normalized Excess Path Error (NEPE) is measured relative to PDM (stepwise), which projects after every inner Langevin update and has NEPE $0$; terminal correction has NEPE $1$. Across budgets, adaptive scheduling remains below periodic correction, showing that defect-aware projection timing is a better use of the same projection budget.}
    \label{fig:pdm-nepe}
    \vspace{-24pt}
\end{wrapfigure}

To understand why scheduling matters, we examine how defect is distributed along the rollout. Fig.~\ref{fig:regime} visualizes defect over time together with correction events for different policies. Defect is highly non-uniform: large excursions are concentrated in specific regions of the trajectory, while many steps incur negligible deviation.

Periodic schedules allocate corrections uniformly, ignoring this structure. In contrast, adaptive schedules concentrate corrections on high-defect regions, avoiding unnecessary projections elsewhere. This behavior is consistent with the budgeted formulation in Sec.~\ref{sec:acs}: defect estimates marginal value, while the $(t,b)$-dependent threshold determines whether that value justifies spending one of the remaining corrections.

This mechanism explains the improvements observed in Fig.~\ref{fig:pathwise} and Tab.~\ref{tab:summary}. By allocating corrections to the small subset of steps that dominate trajectory distortion, adaptive schedules approximate the optimal budget allocation.

\subsection{Generalization to diffusion and planning domains}

\begin{figure}[b]
    \centering
    \includegraphics[width=\linewidth]{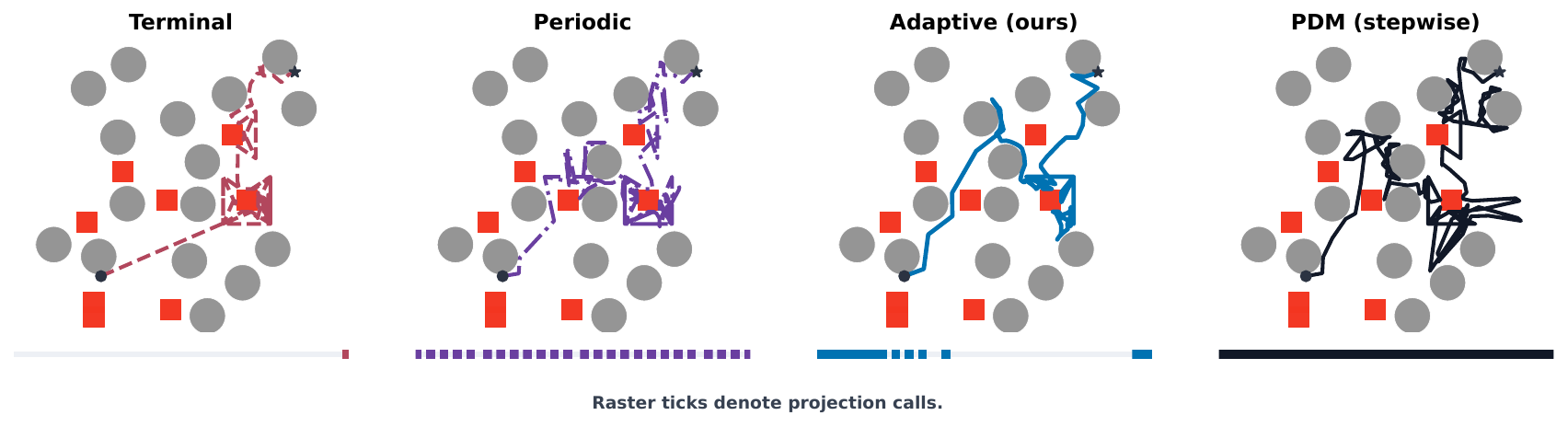}
    \vspace{-12pt}
    \caption{\emph{Projection timing changes constrained diffusion trajectories.} We wrap the Projected Diffusion Models (PDM) trajectory sampler and vary only when its projection operator is applied. Terminal correction delays projection until the end, producing paths that visibly collide with obstacles before being corrected. Periodic correction uses the same projection budget as adaptive but spends it uniformly, leaving avoidable detours and collisions. Adaptive correction spends projections only at high-defect steps, yielding a trajectory closer to PDM (stepwise), which projects after every inner Langevin update. Raster ticks denote projection calls.}
    \label{fig:pdm-qualitative}
    \vspace{-8pt}
\end{figure}

\begin{table}[t]
    \vspace{-4pt}
    \centering
    \small
    \begin{tabular}{lccccc}
        \toprule
        Schedule
        & Cost $B/T$ $\downarrow$
        & NEPE $\downarrow$
        & Benefit recovered $\uparrow$
        & $\Delta$ vs. periodic $\uparrow$ \\
        \midrule
        Periodic
        & 0.25
        & 0.486 $\pm$ 0.009
        & 51.4\%
        & -- \\
        Adaptive (budgeted)
        & 0.25
        & \textbf{0.288 $\pm$ 0.009}
        & \textbf{71.2\%}
        & \textbf{41\%} \\
        \bottomrule
    \end{tabular}
    \vspace{4pt}
    \caption{\emph{Scheduling projection inside Projected Diffusion Models at fixed budget.} Original PDM projects after every inner Langevin update and defines Normalized Excess Path Error (NEPE) $=0$; terminal correction defines NEPE $=1$. At $25\%$ of the projection cost, adaptive scheduling recovers $71.2\%$ of the pathwise benefit of original PDM, compared with $51.4\%$ for periodic correction. Thus, adaptive timing gives a $41\%$ reduction in excess pathwise error over periodic correction at the same budget.}
    \label{tab:pdm-fixed-budget}
\end{table}

Finally, we test whether the same scheduling principle applies inside an existing constrained, diffusion sampler. Projected Diffusion Models (PDM) enforce constraints by applying projection throughout sampling~\citep{christopherConstrainedSynthesisProjected2024, liangSimultaneousMultiRobotMotion2025}. In our terminology, the original PDM sampler is the stepwise baseline, projecting after every inner Langevin update. We keep the trained model, constraint set, and projection operator fixed, and vary only the timing of projection. Rather than introducing a new sampler, we wrap an existing projected diffusion method and replace its all-step projection rule with terminal, periodic, and online budgeted schedules. We measure normalized excess pathwise error (NEPE) relative to the original PDM sampler, so original PDM has NEPE $0$ and terminal correction has NEPE $1$. Lower values therefore indicate closer agreement with the fully projected PDM trajectory. A complete definition of NEPE appears in App.~\ref{app:nepe}.

Fig.~\ref{fig:pdm-qualitative} visualizes representative trajectories under each projection schedule. Terminal correction satisfies the constraint only after the trajectory has already drifted, while periodic correction improves stability by spending projections uniformly. The adaptive scheduler instead allocates the same projection budget according to the realized defect profile, producing trajectories that remain visibly closer to the original PDM sampler while using only sparse projection events.

Fig.~\ref{fig:pdm-nepe} quantifies this behavior across projection budgets. At every tested budget, adaptive scheduling recovers more of the original PDM behavior than periodic correction. The fixed-budget summary in Tab.~\ref{tab:pdm-fixed-budget} shows that at $B/T \approx 0.25$, periodic correction achieves NEPE $0.486 \pm 0.009$, while adaptive scheduling reduces this to $0.288 \pm 0.009$, a $41\%$ reduction in excess pathwise error at the same projection budget. Since original PDM projects after every inner Langevin update, this policy recovers $71.2\%$ of PDM's performance at $25\%$ of the cost.

Thus, even when the projection operator and diffusion model are inherited from a prior constrained diffusion method, projection timing remains a meaningful algorithmic degree of freedom.

\section{Discussion}
\label{sec:discussion}

Projection is usually treated as a feasibility operation: run the sampler, then force the result back onto the constraint set. Our results suggest a different view: in state-dependent generative rollouts, projection is an intervention in the dynamics, and delaying it changes the states from which future updates are computed. Applying it everywhere avoids this drift, but spends correction effort indiscriminately. Correction timing is therefore a budgeted control decision rather than a binary choice between terminal and stepwise projection.

At matched projection budgets, adaptive scheduling improves pathwise fidelity over periodic correction by spending projections on high-defect steps rather than uniformly in time. The gains are largest when defect is concentrated, and smaller when violations are diffuse, which is exactly the regime predicted by the scheduling view. In the PDM experiment, the same idea improves a learned projected
diffusion sampler without changing its model, constraints, or projection operator.

\subsection{Limitations and future work}
\label{sec:limitations}

Our scheduler uses one-step defect as a proxy for downstream trajectory error. This is effective when local constraint violation identifies steps that will perturb future updates, but it can fail when defect is poorly aligned with trajectory distortion or when small violations are dynamically amplified. The method also assumes access to a stable projection operator. If projection is cheap, stepwise correction may be preferable, and if projection is unstable or approximate, both the defect signal and corrected rollout may degrade. Finally, our learned diffusion experiment isolates projection timing by wrapping an existing PDM sampler, but larger models and higher-dimensional constraints may introduce interactions or noise not captured here.

Future work could learn correction value directly, add lookahead or uncertainty estimates, and train generative models with the scheduler in the loop. More broadly, constrained generation should treat feasibility operations as trajectory-level interventions: the path that produces a feasible sample matters.

\bibliographystyle{plainnat}
\bibliography{references}


\appendix

\section{Overview}

This appendix provides additional details for the formulation, implementation, metrics, and experiments in the main paper. Sec.~\ref{app:notation} fixes notation and describes all correction schedules, Sec.~\ref{app:budgeted-policy} gives the full online budgeted scheduler and threshold construction, Sec.~\ref{app:metrics} defines all reported metrics, Sec.~\ref{app:synthetic-domains} describes the controlled manifold experiments, Sec.~\ref{app:pdm} details the Projected Diffusion Models experiment, Sec.~\ref{app:implementation} gives implementation details, Sec.~\ref{app:reproducibility} gives reproducibility notes, and Sec.~\ref{app:diagnostics} gives additional diagnostics and failure modes.

\section{Notation and Correction Schedules}
\label{app:notation}

\paragraph{Generative rollout.}
We consider a generative rollout in an ambient space $\mathcal X \subseteq \mathbb R^d$ with a constraint set $\mathcal M \subset \mathcal X$. A rollout is generated by repeatedly applying an update operator \[
    \tilde x_{t+1} = \Phi_h(x_t), \qquad t = 0,\ldots,T-1,
\] where $h$ denotes the step size or sampler discretization parameter. In diffusion models, $\Phi_h$ may be one reverse diffusion update, one denoising update, or one inner Langevin update, depending on the sampler. In all experiments, the scheduling horizon $T$ counts the number of update locations at which projection could be applied.

\paragraph{Projection or retraction.}
We assume access to a correction map \[
    \Pi : \mathcal X \to \mathcal M,
\] which maps a proposed state back to the constraint set. We use ``projection'' broadly: $\Pi$ may be an exact Euclidean projection, a retraction, a constraint solver, or the projection operator inherited from a projected diffusion method. The scheduler does not require gradients through $\Pi$.

\paragraph{Defect.}
For a proposed update $\tilde x_{t+1}$, the one-step defect is \[
    s_t = d(\tilde x_{t+1}, \mathcal M),
\] where $d$ is a domain-specific constraint violation or distance-to-feasibility score. When a closed-form defect is unavailable, we use the projection residual \[
    d(x,\mathcal M) = \|x-\Pi(x)\|,
\] with the norm chosen to match the state representation.

\paragraph{Correction schedule.}
A correction schedule is a binary policy \[
    \sigma : \{0,\ldots,T-1\} \to \{0,1\},
\] where $\sigma(t)=1$ means projection is applied after update $t$. The corrected rollout is \[
    x_{t+1} = \begin{cases}
        \Pi(\tilde x_{t+1}) & \text{if } \sigma(t)=1,\\
        \tilde x_{t+1} & \text{if } \sigma(t)=0.
    \end{cases}
\]

\paragraph{Standard schedules.}
The main paper compares four correction schedules: \begin{itemize}
    \item \textbf{Terminal correction:} no intermediate projection is applied, and projection is applied only at the end of the rollout.
    \item \textbf{Stepwise correction:} projection is applied after every update.
    \item \textbf{Periodic correction:} projection is applied uniformly in time subject to a fixed projection budget.
    \item \textbf{Adaptive budgeted correction:} projection is applied online according to defect, time, and remaining budget.
\end{itemize}

\paragraph{Projection budget.}
For a rollout of length $T$, a budget $B$ permits at most $B$ projection calls. We report budget as the fraction \[
    B/T.
\] Periodic and adaptive schedules are compared at matched $B/T$. Stepwise correction has $B/T=1$. Terminal correction has only a final correction; when normalized against stepwise cost, its intermediate projection budget is effectively zero.

\section{Online Budgeted Scheduling}
\label{app:budgeted-policy}

\subsection{Ideal budgeted allocation}

The scheduling problem can be written as a finite-budget allocation problem. Let $V_t(x_t)$ denote the downstream value of correcting at step $t$, i.e. the expected reduction in future trajectory error if projection is applied at the current proposal. The ideal schedule solves \[
    \max_{\sigma}
    \sum_{t=0}^{T-1} \sigma(t)V_t(x_t)
    \qquad
    \text{s.t.}
    \qquad
    \sum_{t=0}^{T-1}\sigma(t) \le B,
\] where $\sigma(t)\in\{0,1\}$.

This formulation is conceptually useful but intractable in practice, as estimating $V_t(x_t)$ requires knowing how a correction changes future stochastic updates, which depends on the remaining sampler trajectory. We therefore approximate marginal value using the observable one-step defect $s_t$.

\subsection{Budget-aware thresholds}

The online budgeted scheduler uses thresholds indexed by time and remaining budget: \[
    \lambda_{t,b},
    \qquad
    t=0,\ldots,T-1,\quad b=0,\ldots,B.
\] The threshold $\lambda_{t,b}$ is interpreted as the marginal price of spending one correction at time $t$ with $b$ corrections remaining. Given a proposed update $\tilde x_{t+1}$ and defect $s_t$, the policy is \[
    x_{t+1} =
    \begin{cases}
        \Pi(\tilde x_{t+1})
        & \text{if } b_t>0 \text{ and } s_t \ge \lambda_{t,b_t},\\
        \tilde x_{t+1}
        & \text{otherwise.}
    \end{cases}
\] If projection is applied, then $b_{t+1}=b_t-1$; otherwise $b_{t+1}=b_t$.

The dependence on both $t$ and $b$ is important. Early in the rollout, a moderate defect may not justify spending a scarce projection if many future opportunities remain. Late in the rollout, the same defect may be worth correcting because unused budget has less future value. The threshold surface $\lambda_{t,b}$ captures this online tradeoff.

\subsection{Empirical threshold construction}
\label{app:threshold-construction}

We estimate $\lambda_{t,b}$ on held-out calibration rollouts disjoint from evaluation seeds. The calibration procedure collects defect traces from rollouts without intermediate adaptive correction. Let \[
    \mathcal S_t = \{s_{i,t}\}_{i=1}^{N_{\mathrm{cal}}}
\] denote the empirical distribution of defects at time $t$ across calibration rollouts. A simple budget-aware threshold can be constructed by considering the future defect pool \[
    \mathcal S_{\ge t} = \{s_{i,u}: i=1,\ldots,N_{\mathrm{cal}},\ u=t,\ldots,T-1\}.
\] For remaining budget $b$, we set \[
    \lambda_{t,b}
    =
    Q_{1 - b/(T-t)}(\mathcal S_{\ge t}),
\] where $Q_q$ is the empirical $q$-quantile. The convention is: \[
    \lambda_{t,0}=+\infty,
    \qquad
    \lambda_{t,b}=-\infty \text{ if } b\ge T-t.
\] Thus, when no budget remains, correction is impossible; when enough budget remains to correct every future step, the scheduler corrects all remaining proposals.

This rule estimates the defect level above which a future proposal belongs to the top $b$ remaining events. It is online, model-agnostic, and uses only held-out defect statistics.

\subsection{Budget compliance}

Adaptive schedules are evaluated against periodic schedules with the same nominal budget $B/T$. In implementations where the threshold rule may underspend, we record the achieved budget \[
    \widehat B/T
    =
    \frac{1}{T}
    \sum_{t=0}^{T-1}\sigma(t).
\]
Main results use settings in which adaptive and periodic budgets are matched up to numerical tolerance. Tables report the target budget, and logs record achieved budget means and standard errors.

\subsection{Algorithm}

\begin{algorithm}[h]
    \caption{Online budgeted adaptive correction}
        \label{alg:appendix-adaptive}
        \begin{algorithmic}[1]
        \Require rollout length $T$, budget $B$, thresholds $\{\lambda_{t,b}\}$, update rule $\Phi_h$, projection $\Pi$, defect $d$
        \State $b \gets B$
        \For{$t=0,\ldots,T-1$}
            \State $\tilde x_{t+1} \gets \Phi_h(x_t)$
            \State $s_t \gets d(\tilde x_{t+1},\mathcal M)$
            \If{$b>0$ and $s_t \ge \lambda_{t,b}$}
                \State $x_{t+1} \gets \Pi(\tilde x_{t+1})$
                \State $b \gets b-1$
            \Else
                \State $x_{t+1} \gets \tilde x_{t+1}$
            \EndIf
        \EndFor
    \end{algorithmic}
\end{algorithm}

\section{Metrics}
\label{app:metrics}

\subsection{Endpoint distance to stepwise}

Endpoint distance measures how far a method's final sample deviates from the stepwise reference: \[
    E_{\mathrm{end}}(\sigma)
    =
    \rho(x_T^\sigma, x_T^{\mathrm{step}}),
\] where $\rho$ is the domain-appropriate state distance. For Euclidean states, $\rho$ is the Euclidean norm, and for manifold-valued states, $\rho$ is the corresponding geodesic or product distance. In all cases, endpoint distance is computed between matched rollouts sharing the same initial condition and stochastic seed.

Endpoint distance isolates the distributional shift induced by delayed correction. A terminally projected sample may be feasible, but if it differs substantially from the stepwise endpoint, then final feasibility did not recover the constrained rollout.

\subsection{Pathwise error}

We use pathwise error to measure trajectory-level deviation. In the synthetic experiments, the primary pathwise score is cumulative constraint defect: \[
    E_{\mathrm{path}}(\sigma)
    =
    \sum_{t=1}^{T}
    d(x_t^\sigma,\mathcal M).
\]

For comparisons to a stepwise reference trajectory, we also compute state-space pathwise deviation: \[
    E_{\mathrm{state}}(\sigma)
    =
    \sum_{t=1}^{T}
    \rho(x_t^\sigma, x_t^{\mathrm{step}}).
\]
The main paper reports normalized excess pathwise error using the pathwise score appropriate to each experiment.

\subsection{Normalized excess pathwise error}
\label{app:nepe}

Normalized excess pathwise error (NEPE) is defined between stepwise correction and terminal correction: \[
    \mathrm{NEPE}(\sigma)
    =
    \frac{
        E_{\mathrm{path}}(\sigma)
        -
        E_{\mathrm{path}}(\mathrm{stepwise})
    }{
        E_{\mathrm{path}}(\mathrm{terminal})
        -
        E_{\mathrm{path}}(\mathrm{stepwise})
    }.
\]
Thus stepwise correction has NEPE $0$ and terminal correction has NEPE $1$. Lower values are better. When the denominator is below a numerical tolerance, the rollout is marked degenerate and excluded from normalized summaries. Raw pathwise values are retained in logs.

\subsection{Improvement percentages}

For a metric $m$ where lower is better, the adaptive improvement over periodic is
\[
    \Delta_m
    =
    \frac{
        m_{\mathrm{periodic}} - m_{\mathrm{adaptive}}
    }{
        m_{\mathrm{periodic}}
    }.
\]
Positive values indicate adaptive improvement. Negative values indicate periodic is better. Tables compute $\Delta_m$ from unrounded paired values.

For PDM, we also report the fraction of full correction benefit recovered: \[
    \mathrm{Benefit}(\sigma)
    =
    1-\mathrm{NEPE}(\sigma).
\]

\subsection{Win rates}

Win rates compare adaptive and periodic schedules at matched domain, seed, and projection budget. For metric $m$, the adaptive win indicator is \[
    \mathbf 1[m_{\mathrm{adaptive}} < m_{\mathrm{periodic}}].
\]

The reported win rate is the empirical mean of this indicator across paired comparisons. Uncertainty is reported as binomial standard error: \[
    \mathrm{SE}
    =
    \sqrt{\frac{\hat p(1-\hat p)}{N}},
\] where $\hat p$ is the observed win rate and $N$ is the number of paired comparisons.

\section{Synthetic Manifold Experiments}
\label{app:synthetic-domains}


\subsection{Overview}

The synthetic experiments are designed to isolate correction timing under controlled geometry. Each domain specifies:
\begin{itemize}
    \item a constraint set $\mathcal M$;
    \item an ambient update rule $\Phi_h$;
    \item a projection or retraction $\Pi$;
    \item a defect function $d(x,\mathcal M)$;
    \item a distance $\rho$ used for endpoint and state-space errors.
\end{itemize}

We evaluate six domains:
\[
    \mathrm{SO}(3),\quad
    \mathrm{SE}(3),\quad
    \mathrm{Terrain},\quad
    \mathrm{SO}(3)\text{-Impulse},\quad
    \mathrm{SE}(3)\text{-Lever},\quad
    \mathrm{Terrain}\text{-Ridge}.
\]

The first three provide smooth baseline settings. The latter three introduce localized high-defect events, producing the heterogeneous defect profiles for which adaptive scheduling is designed.

\subsection{$\mathrm{SO}(3)$}

States are represented as matrices in $\mathbb R^{3\times 3}$ with the constraint set \[
    \mathcal M = \mathrm{SO}(3)
    =
    \{R\in\mathbb R^{3\times 3}: R^\top R=I,\ \det(R)=1\}.
\]

Projection is implemented by polar decomposition. Given an ambient matrix $A$, compute \[
    A = U\Sigma V^\top,
\] and set \[
    \Pi(A)=UV^\top,
\] with the determinant corrected if necessary. The defect is the orthogonality residual \[
    d(A,\mathrm{SO}(3))
    =
    \|A^\top A-I\|_F
    +
    |\det(A)-1|.
\]

Endpoint distances use the geodesic rotation distance \[
    \rho(R_1,R_2)
    =
    \|\log(R_1^\top R_2)\|_F/\sqrt{2}.
\]

\subsection{$\mathrm{SE}(3)$}

States consist of a rotation and translation pair $(R,p)$, with \[
    R\in \mathrm{SO}(3),\qquad p\in\mathbb R^3.
\]

Projection applies the $\mathrm{SO}(3)$ projection to the rotational block and leaves translation unchanged unless the domain-specific constraint requires translation correction. The defect combines rotational feasibility and translational constraint violation: \[
    d((A,p),\mathrm{SE}(3))
    =
    d(A,\mathrm{SO}(3)) + \alpha d_{\mathrm{trans}}(p).
\]

Endpoint distance is the weighted product metric \[
    \rho((R_1,p_1),(R_2,p_2))
    =
    \rho_{\mathrm{SO}(3)}(R_1,R_2)
    +
    \alpha \|p_1-p_2\|_2.
\]

\subsection{Terrain}

The terrain domain uses a graph-like constraint manifold \[
    \mathcal M
    =
    \{(u,v,z): z=f(u,v)\},
\] where $f$ is a smooth terrain height field. Projection maps an ambient point $(u,v,z)$ to \[
    \Pi(u,v,z)=(u,v,f(u,v)).
\]

The defect is vertical deviation: \[
    d((u,v,z),\mathcal M)=|z-f(u,v)|.
\]

Endpoint and pathwise distances are computed in ambient Euclidean coordinates.

\subsection{Impulse, lever, and ridge variants}

The volatile variants introduce localized regions where the ambient dynamics produce larger mismatch with the constraint set. These variants are not separate methods; their purpose is to create rollouts in which defect mass is concentrated in a small subset of steps.

\paragraph{$\mathrm{SO}(3)$-Impulse.}
The impulse variant adds a localized ambient perturbation to the rotation update, producing short bursts of large orthogonality defect.

\paragraph{$\mathrm{SE}(3)$-Lever.}
The lever variant couples rotational and translational errors so that small rotational drift can induce amplified endpoint displacement. This tests whether adaptive correction can identify geometrically consequential errors rather than merely large Euclidean deviations.

\paragraph{Terrain-Ridge.}
The ridge terrain adds high-curvature regions to the height field. Rollouts crossing the ridge produce localized projection--dynamics mismatch, while flatter regions remain low-defect.

\subsection{Defect concentration}

To quantify how concentrated defect is along a rollout, we compute top-$q$ defect mass. Let $s_1,\ldots,s_T$ be the defect sequence and let $I_q$ index the largest $\lceil qT\rceil$ defect values. The top-$q$ mass is \[
    C_q
    =
    \frac{
        \sum_{t\in I_q} s_t
    }{
        \sum_{t=1}^{T}s_t
    }.
\]

In the main experiments, $q=0.2$ unless otherwise stated. Large $C_q$ indicates that a small fraction of steps accounts for most of the defect. This is the regime where adaptive scheduling should have the largest advantage over periodic correction.

\subsection{Additional Results}
\label{app:additional-results}

\begin{figure}
    \centering
    \includegraphics[width=\linewidth]{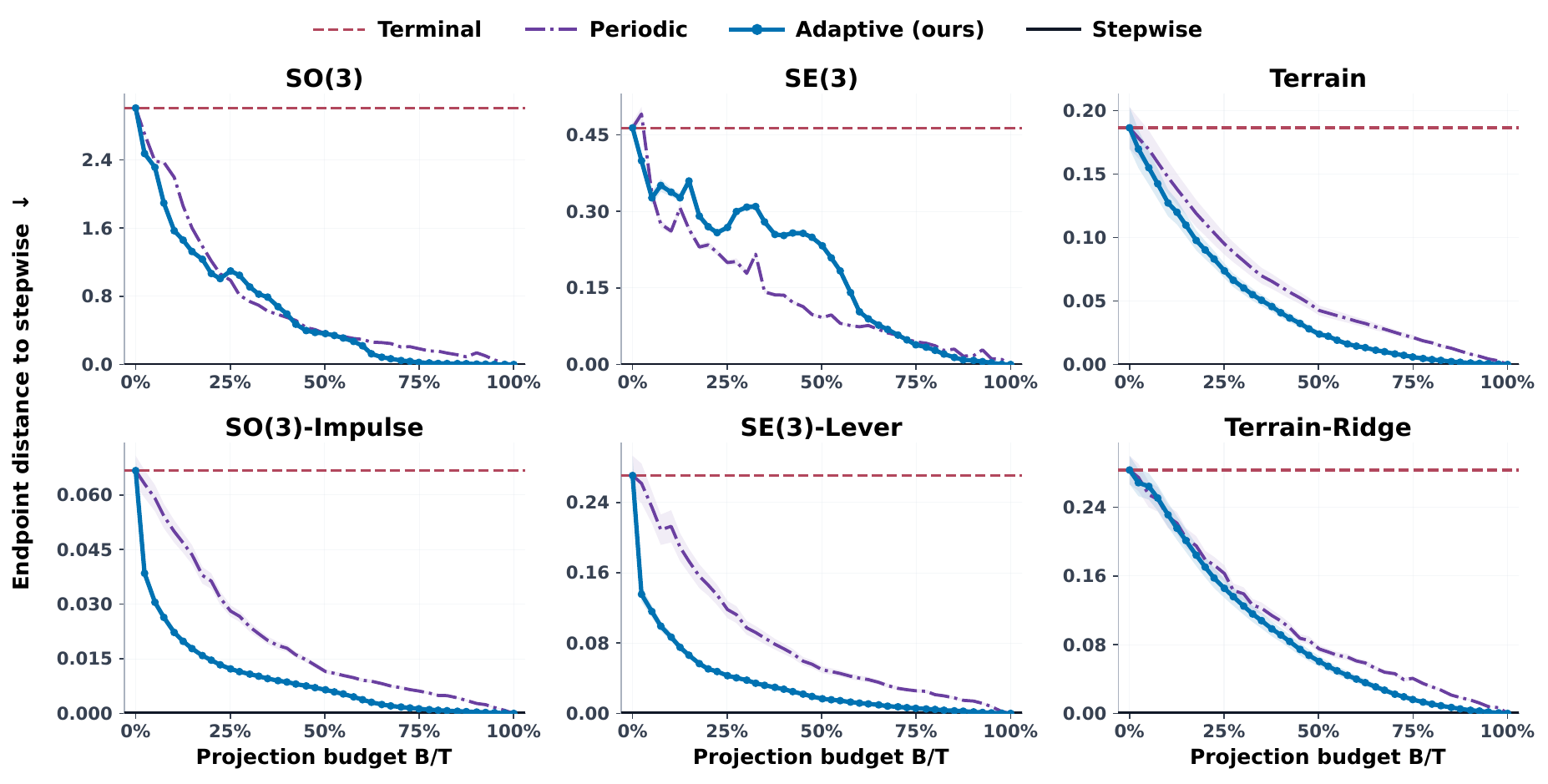}
    \caption{\emph{Endpoint distance to the stepwise constrained reference.} Delayed correction can change the final sample even when a terminal projection restores feasibility. Adaptive scheduling often reduces this endpoint shift at fixed projection budget, especially in volatile domains, but endpoint gains are more geometry-dependent than pathwise gains. This complements the main Normalized Excess Path Error results in Fig.~\ref{fig:pathwise}.}
    \label{fig:endpoint}
\end{figure}

\begin{table}[b]
    \vspace{-8pt}
    \centering
    \begin{tabular}{lcccc}
        \toprule
        Domain & Endpoint win $\uparrow$ & Pathwise win $\uparrow$ & Median $\Delta$ NEPE $\uparrow$ & $N$ \\
        \midrule
        SO(3) & \textbf{76\% $\pm$ 2\%} & \textbf{90\% $\pm$ 2\%} & \textbf{85\%} & 336 \\
        SE(3) & 59\% $\pm$ 3\% & \textbf{90\% $\pm$ 2\%} & \textbf{31\%} & 336 \\
        Terrain & 49\% $\pm$ 3\% & \textbf{85\% $\pm$ 2\%} & \textbf{52\%} & 336 \\
        SO(3)-Impulse & \textbf{90\% $\pm$ 2\%} & \textbf{90\% $\pm$ 2\%} & \textbf{67\%} & 336 \\
        SE(3)-Lever & \textbf{79\% $\pm$ 2\%} & \textbf{86\% $\pm$ 2\%} & \textbf{48\%} & 336 \\
        Terrain-Ridge & 43\% $\pm$ 3\% & \textbf{88\% $\pm$ 2\%} & \textbf{48\%} & 336 \\
        \bottomrule
    \end{tabular}
    \vspace{4pt}
    \caption{\emph{Adaptive scheduling is reliable on paired budget-matched comparisons.} Each comparison matches domain, seed, and projection budget. Endpoint win rate reports how often adaptive ends closer to the stepwise reference than periodic; pathwise win rate reports how often adaptive obtains lower Normalized Excess Path Error. Adaptive is especially reliable on the pathwise metric, winning $85$--$90\%$ of paired comparisons across domains, while endpoint wins vary with geometry.}
    \label{tab:winrate}
\end{table}

Fig.~\ref{fig:endpoint} provides visualization of endpoint distance from stepwise to periodic and adaptive policies alongside the metrics in Tab.~\ref{tab:summary}. Tab.~\ref{tab:winrate} provides paired win-rate diagnostics to accompany the rollouts reported in Tab.~\ref{tab:summary}.

\section{Projected Diffusion Models Experiment}
\label{app:pdm}

\subsection{Purpose}

The PDM experiment tests whether correction scheduling can wrap an existing constrained diffusion sampler. Unlike the controlled synthetic domains, this experiment does not introduce a new generative model or projection operator: we inherit the model, constraints, and projection machinery from Projected Diffusion Models (PDM), and vary only when projection is applied~\citep{christopherConstrainedSynthesisProjected2024, liangSimultaneousMultiRobotMotion2025}.

\subsection{Original PDM as stepwise projection}

PDM applies projection throughout sampling. In the trajectory experiment used here, projection is applied after every inner Langevin update. Therefore, in our terminology, original PDM is the stepwise baseline: \[
    \sigma_{\mathrm{PDM}}(t)=1
    \qquad
    \forall t.
\]

The scheduling horizon $T$ counts inner Langevin updates, not just outer diffusion noise levels. This is important for cost accounting: if the sampler uses $K$ outer noise levels and $L$ inner Langevin steps per level, then \[
    T = K L.
\]

A budget $B/T=0.25$ therefore means projection is applied to one quarter of all inner Langevin update locations.

\subsection{Wrapped schedules}

We evaluate four variants: \begin{itemize}
    \item \textbf{Original PDM / stepwise:} project after every inner Langevin update.
    \item \textbf{Terminal:} run the sampler without intermediate projection and apply projection at the end.
    \item \textbf{Periodic:} apply projection uniformly over inner Langevin updates under budget $B$.
    \item \textbf{Adaptive budgeted:} apply projection when $s_t\ge \lambda_{t,b_t}$ and budget remains.
\end{itemize}

All variants use the same trained PDM model, the same constraints, and the same projection operator.

\subsection{Defect for PDM}

When PDM exposes an explicit constraint violation, we use that violation as the defect. Otherwise, we define defect by projection residual: \[
    s_t
    =
    \|\tilde x_{t+1}-\Pi(\tilde x_{t+1})\|.
\]

This measures how far the proposed state lies from the constraint set under the same projection operator used by PDM.

\subsection{Metrics}

PDM NEPE is normalized between original PDM and terminal correction: \[
    \mathrm{NEPE}(\sigma)
    =
    \frac{
        E_{\mathrm{path}}(\sigma)-E_{\mathrm{path}}(\mathrm{PDM})
    }{
        E_{\mathrm{path}}(\mathrm{terminal})-E_{\mathrm{path}}(\mathrm{PDM})
    }.
\]

Thus, original PDM has NEPE $0$ and terminal correction has NEPE $1$. Lower values indicate closer agreement with the fully projected PDM trajectory.

\subsection{Fixed-budget result}

At $B/T\approx0.25$, periodic correction achieves NEPE $0.486 \pm 0.009$, while adaptive budgeted correction achieves NEPE $0.288 \pm 0.009$. The improvement over periodic is $41\%$. Since original PDM uses projection at every inner Langevin update, this operating point uses only $25\%$ of the original projection calls, saving $75\%$ of projection calls. In absolute terms, adaptive scheduling recovers $71.2\%$ of the full PDM correction benefit at one quarter of the projection cost.

\section{Implementation Details}
\label{app:implementation}

\subsection{Random seeds and paired evaluation}

All comparisons are paired by seed. For each initial condition and random seed, we run terminal, periodic, adaptive, and stepwise schedules under the same underlying stochastic updates whenever possible. This reduces variance and ensures that differences are attributable to projection timing rather than different noise realizations.

\subsection{Calibration/evaluation split}

Thresholds are estimated using held-out calibration rollouts, and evaluation rollouts are disjoint from calibration rollouts. Synthetic and PDM experiments each use separate calibration and evaluation seeds to prevent the adaptive scheduler from selecting thresholds based on evaluation trajectories.

\subsection{Budget grid}

Experiments evaluate a grid of projection budgets from $0.00$ to $1.00$ with a step size of $0.05$. Tables report a representative operating point at $B/T\approx0.25$.

\subsection{Uncertainty}
\label{app:uncertainty}

Curves report means over seeds. Shaded regions, when shown, indicate standard error of the mean. Tables report mean $\pm$ standard error unless otherwise stated. Win-rate uncertainties use binomial standard error.

\subsection{Numerical safeguards}

Normalized metrics are not reported when the normalization denominator is below tolerance. Specifically, if \[
    E_{\mathrm{path}}(\mathrm{terminal})
    -
    E_{\mathrm{path}}(\mathrm{stepwise})
    < \epsilon,
\] the corresponding NEPE value is marked degenerate. This prevents near-identical terminal and stepwise runs from producing unstable normalized values.

\subsection{Projection cost}

The primary cost metric is the number of projection calls. This is the relevant cost for settings where projection or constraint solving is expensive relative to a model update. We do not claim that projection count exactly equals wall-clock time in every domain; rather, it gives a model-independent accounting of constraint enforcement effort. In PDM, where projection is applied after inner Langevin updates, cost is counted at the inner-update level.

\section{Reproducibility Notes}
\label{app:reproducibility}

A full evaluation run consists of:
\begin{enumerate}
    \item generating calibration rollouts;
    \item estimating $\lambda_{t,b}$;
    \item running paired evaluation rollouts for all schedules;
    \item aggregating endpoint and pathwise metrics;
    \item producing figures and tables.
\end{enumerate}

All full paper evaluations run on a single NVIDIA A100 GPU via Modal in fewer than $8$ hours, with lower-fidelity and preliminary evaluations running within $24$ hours on an Apple M1 (16GB).

\section{Additional Diagnostics}
\label{app:diagnostics}

\subsection{Budget usage}

For each method, domain, and target budget, we log: \[
    \widehat B/T
    =
    \frac{1}{T}
    \sum_{t=0}^{T-1}\sigma(t).
\]
Periodic schedules use the requested budget by construction up to rounding. Adaptive schedules are checked to ensure that achieved budgets match periodic budgets within tolerance.

\subsection{Degenerate regimes}

Some domains or sampler settings may produce little difference between terminal and stepwise correction. In such cases, NEPE becomes unstable because the denominator is small. These regimes indicate that projection timing is not consequential under that sampler and constraint configuration. We report raw pathwise errors in diagnostics and exclude degenerate normalized values.

\subsection{Failure modes}
\label{app:failure-modes}

The method can fail or underperform when:
\begin{itemize}
    \item defect is poorly aligned with downstream trajectory distortion;
    \item the projection operator is unstable or discontinuous;
    \item constraint violation is diffuse and uniform, making periodic correction competitive;
    \item the adaptive threshold surface is poorly calibrated;
    \item projection cost is negligible, making stepwise correction preferable.
\end{itemize}

These cases are consistent with the scheduling interpretation, that adaptive allocation is most useful when projection is costly and defect is concentrated.



\end{document}